\documentclass[12pt,letter]{article}
\usepackage[pdftex]{graphicx} 
\usepackage{array} 
\usepackage{graphicx,color,enumerate} 
\usepackage[square,sort&compress,numbers]{natbib}

\usepackage{epsfig}

\usepackage{geometry}
\usepackage{amsmath,amsfonts,amsthm}

\usepackage{multirow}
\newcommand{\CC}{\bf{C}}
\newcommand{\JJ}{\mathcal{J}}

\begin{document}
%\singlespacing
%
% paper title
% can use linebreaks \\ within to get better formatting as desired
\title{Fast Imbalanced Classification of Healthcare Data with Missing Values}

% author names and affiliations
% use a multiple column layout for up to three different
% affiliations
%\author{
%\IEEEauthorblockN{Talayeh Razzaghi}
%\IEEEauthorblockA{School of Computing, Clemson University\\
%Clemson, SC 29630\\
%Email:  trazzag@clemson.edu}
%\and
%\IEEEauthorblockN{Oleg Roderick}
%\IEEEauthorblockA{Geisinger Health System \\
% Danville, PA 17822\\
%Email: oeroderick@geisinger.edu}
%\and
%\IEEEauthorblockN{Ilya Safro}
%\IEEEauthorblockA{School of Computing, Clemson University\\
%Clemson, SC 29630\\
%Email:  isafro@clemson.edu}
%\and
%\IEEEauthorblockN{Nick Marko}
%\IEEEauthorblockA{Geisinger Health System \\
% Danville, PA 17822\\
%Email: nmarko@geisinger.edu}
%}

% conference papers do not typically use \thanks and this command
% is locked out in conference mode. If really needed, such as for
% the acknowledgment of grants, issue a \IEEEoverridecommandlockouts
% after \documentclass

% for over three affiliations, or if they all won't fit within the width
% of the page, use this alternative format:
% 
\author{Talayeh Razzaghi\\
School of Computing\\
Clemson Univeristy\\
Clemson, SC 29634\\ 
Email: trazzag@clemson.edu
\and
Oleg Roderick\\
Geisinger Health System\\
Danville, PA 17822\\
Email: oroderick@geisinger.edu
\and
Ilya Safro\\
School of Computing\\
Clemson Univeristy\\
Clemson, SC 29634\\ 
Email: isafro@clemson.edu
\and 
Nick Marko\\
Geisinger Health System\\
Danville, PA 17822\\
Email: nmarko@geisinger.edu
}

%\renewcommand{\thesubsection}{\arabic{subsection}}

% use for special paper notices
%\IEEEspecialpapernotice{(Invited Paper)}

% make the title area
\maketitle

\begin{abstract}
%\boldmath
In medical domain, data features often contain missing values. This can create serious bias in
the predictive modeling. Typical standard data mining methods often produce poor performance measures.
In this paper, we propose a new method to simultaneously classify large datasets and reduce the effects
of missing values. The proposed method is based on a multilevel framework of the cost-sensitive SVM and
the expected maximization imputation method for missing values, which relies on iterated regression
analyses. We compare classification results of multilevel SVM-based algorithms on public benchmark datasets
with imbalanced classes and missing values as well as real data in health applications, and show that our multilevel
SVM-based method produces fast, and more accurate and robust classification results.
\end{abstract}

% IEEEtran.cls defaults to using nonbold math in the Abstract.
% This preserves the distinction between vectors and scalars. However,
% if the conference you are submitting to favors bold math in the abstract,
% then you can use LaTeX's standard command \boldmath at the very start
% of the abstract to achieve this. Many IEEE journals/conferences frown on
% math in the abstract anyway.

% no keywords

\section{The role of predictive modeling in healthcare}\label{sec:intro}
Modern healthcare can be characterized as \textit{evidence-driven} and \textit{model-assisted} \cite{foldy2014national}. In an ideal situation, every decision in the clinical environment should be supported by a statistical model predicting risks and positive outcomes. This model may have a form of a simplified risk-assessment formula \cite{haas2013risk}, or a sophisticated machine learning tool \cite{plis2014machine}, \cite{woolery1994machine}. In either case, it is based on a query of relevant clinical and operational history. 

In practice, comprehensive medical information is stored in multiple databases, with different formats and rules of access. Due to considerations of patient privacy, and the proprietary nature of electronic medical records \cite{larson2004survey}, the databases cannot be queried continuously. Every instance of data acquisition and integration is a separate effort that is cost-effective only when the resulting predictive model shows high quality. Thus, progress in evidence-driven healthcare depends on how well state-of-the art algorithms of machine learning are adapted to clinical data.

We note that classical computer science issues, such as scalability, or convergence rate are rarely a major issue for healthcare applications. Instead, an algorithm is ranked based on its ability to process raw medical data, with such problematic features as sparsity, missing entries, noise and imbalanced outputs. Because of the encounter nature of patient-provider interaction, medical data is inherently sparse: when a clinical encounter occurs, the number of and contents of labels attached to it vary widely \cite{snomed2011systematized}; outside of an encounter, the state of the patient is unknown. The outcomes of interest in classification problems are imbalanced, because, as a rule, healthcare analytics is motivated by rare events such as healthcare emergencies, severe chronic conditions, gaps and bottlenecks in access to care. 
The extent to which medical data is problematic may not be obvious from the perspective of a local healthcare provider (such as a single doctor); by definition they have access to all knowledge they ever use. We view this paper as a short, high-level primer on using advanced methods of machine learning to overcome the difficulties that emerge after multiple datasets are integrated for analysis and prediction.

This work was prompted by several projects completed with the Division of Applied Research and Clinical Informatics, Dept. of Data Science; Geisinger Health System. The routine activity of Data Science consists of medium-scope predictive projects on a combination of patient biometrics, pathology lab results, clinical encounter data, medical insurance data (available directly from Geisinger Health Plan) and externally assigned aggregate metrics for patients' general lifestyle risks, compliance with treatment regime, and loyalty to a particular provider. 

For the first motivating example (Example 1), we use our 2014 feasibility study \cite{RoderickMEDAI} of merging insurance information (6 aggregate features, based on the history of claims and payments) together with clinical encounter information (10-20 features chosen by hand from patient biometrics, medications and diagnostic codes). The goal of the initial study was to predict the financial risk for a particular patient (a common metric in insurance practice, derived as a ratio of individual expenses and average expenses for a large demographic group). Furthermore, we wanted to see how addition of the clinical information changes the predictive power of the model, thus making a case for existence of high-risk patients that are invisible to claims-based analysis. We used a standard clustering technique, k-nearest neighbors with empirically selected weighting, to achieve the basic results.

For Example 2, we use our preliminary investigation of patients' response to public outreach \cite{RoderickFLU}, such as annual flu awareness campaigns. We included basic demographic and clinical information on patients targeted by 35 such campaigns into the model predicting whether a given patient is likely to respond to the reminder, or to choose not to get vaccinated, or use a different provider. Again, our core predictive model was standard: logistic regression with empirically selected weighting of training data.

We now pose a question: how much more effective would the predictive models be in each case with the use of an advanced machine learning algorithm developed with awareness of sparsity and class skewness (imbalance) in data?

\section{Support Vector Machine Algorithms for Medical Data}
Given a training set $\mathcal{J}=\{(x_i, y_i)\}_{i=1}^l$, that is a set of data points with known labels, where $(x_i, y_i)~\in~\mathbb{R}^{n+1}$, and  $l$ and $n$ are the numbers of data points and features, respectively, and $y_i \in \{-1,1\}$ denotes the class label for each data point $i$ in $\mathcal{J}$. We denote by $\CC^-$ and $\CC^+$, the "majority'' (points with $y_i=+1$) and ``minority'' (points with $y_i=-1$) classes respectively such that $\mathcal{J}=\CC^+ \cup \CC^-$. 

\subsection{Support Vector Machines}
The support vector machine (SVM) solves the following max-margin problem:
\begin{subequations}\label{softmarginSVM}
\begin{align}  
              \min &\ \ {\frac{1}{2}\ \| w \|^2+C \sum_{i=1}^{l}\xi_i} \\
               \text{s.t.}&\ \               y_i(w^T\phi(x_i)+b)\geq1-\xi_i &  \;i = 1, \ldots, l\\
            & \ \  \xi_i\geq0  &     \;i = 1, \ldots, l
\end{align}
\end{subequations}
\noindent where the optimal margin is defined by the parameters $w$ and $b$. The training data points $x_i$ are mapped into a higher dimensional space through  function $\phi: \mathbb{R}^{n} \to \mathbb{R}^{m}$  ($m\geq n$). The misclassified points are penalized using the term slack variables $\xi_i$ ($i \in \{1, \ldots, l\}$) and the parameter $C>0$ controls the magnitude of penalization. Hence, this formulation is called as {\em soft margin} SVM.  
The primal formulation is usually transformed to the Lagrangian dual problem using different algorithms. One of the most popular is the sequential minimal optimization (SMO) which is implemented in the LIBSVM tool \cite{chang2011libsvm}, since it is fast and yields reliable convergence.

\subsection{Weighted Support Vector Machines}

A cost-sensitive extension of SVM, developed to cope with imbalanced data, is known as \emph{weighted} SVM (WSVM) \cite{veropoulos1999controlling}. The main idea is to consider weighting scheme in learning such that the WSVM algorithm builds the decision hyperplane based on the relative contribution of data points in training. In contrast to the standard SVM, the penalization costs are different for the positive ($C^+$) and negative ($C^-$) classes: 
\begin{subequations}\label{eq:wsvm}
\begin{align}  
              \min &\ \ {\frac{1}{2}\ \| w \|^2+C^+ \sum_{\{i|y_i=+1\}}^{n_+}\xi_i+C^-\sum_{\{j|y_j=-1\}}^{n_-}\xi_j}\\
              \textrm{s.t.}&\ \  y_i(w^T\phi(x_i)+b)\geq 1-\xi_i \hspace{20pt}  i = 1, \ldots, l\\
&\ \ \xi_i\geq0 \hspace{100pt}  i = 1, \ldots, l
\end{align}
\end{subequations}

\noindent The formulations (\ref{softmarginSVM}) and (\ref{eq:wsvm}) are solved through the Karush-Kuhn-Tucker conditions.
%The kernel function $k(\bx_i,\bx_j) = {\phi(x_i)}^T \phi(x_j)$ is used to measure the similarity between each pair of points $x_i$ and $x_j$. 
The Gaussian kernel function (radial basis function, RBF) is used in the dual formulation of (W)SVMs since this kernel function usually results into superior performance for many classification problems \cite{tay2001application,xanthopoulos2014weighted}. 
%%The RBF kernel is defined as,
%\begin{equation}
%       k(\bx_i,\bx_j)=\exp(-\gamma \|\bx_i-\bx_j\|^2),  \gamma \geq 0.
% \end{equation} 
Parameter tuning is required to set optimal or near optimal $C$, $C^+$, $C^-$, and kernel function parameters (e.g. bandwidth parameter for RBF kernel function) to achieve good results for (W)SVM. This process becomes problematic and time-consuming particularly when the size of data is very large. Hence we aim to develop an efficient and effective classification method, called the Multilevel (W)SVM, that is scalable and works with imbalanced healthcare data. 

%$\gamma$ 

\subsection{Multilevel Support Vector Machines}
The proposed algorithm belongs to the family of multilevel optimization strategies \cite{brandt:optstrat} whose goal is to approximate the system at multiple scales of coarseness and to obtain a final solution by combining the information from different scales. The multilevel framework for SVM \cite{fastML} scales efficiently for large classification problems whose hierarchy of coarser representations is constructed based on the approximated $k$-nearest neighbors graphs (A$k$NN). This method consists of three main phases:
\begin{itemize}
   \item {\bf The coarsening phase.} A gradual coarsening of the training set is constructed using fast point selection method \cite{saaddim} in A$k$NN graph. However, we found that ensuring a uniform coverage of the points can lead to much better results than finding an independent set of points (nodes in A$k$NN) as was suggested in \cite{saaddim}. Thus, we extended the set of coarse points by setting a parameter for the minimum number of points that in our experiments was set to 50\% of the fine data points.
   
   \item {\bf Supervised support vector initial learning.} After the hierarchy is created, the support vectors learning is performed at the coarsest level, where the number of data points is sufficiently small.
      \item {\bf The uncoarsening phase.} Support vectors, and classifier are projected throughout the hierarchy from the coarsest to the finest levels. At each level, the solution to the current fine level is updated and optimized based on the solution of the previous coarse level. The locally optimal support vectors are obtained by gradual refinement of the projected support vectors from the coarse level. 
\end{itemize}
For imbalanced data, the WSVM can easily be adopted as the base classifier for multilevel framework (MLWSVM). The regular SVM does not  perform well on imbalanced data because it tends to train models with respect to the majority class and technically ignores the minority class. However, the effect of imbalanced issue decreases while using multilevel framework since we prevent creating very small coarse sets for the minority class even if the majority class can still be coarsened.

Often, methods for imbalanced classification demonstrate poor performance on data with missing values (such as \cite{farhangfar2008impact}) that is a frequent situation in healthcare data. Therefore, we apply imputation methods prior the classification model. Such imputation methods have been well studied in  statistical analysis and machine learning domains \cite{ghannad2010selection,little2002statistical,schafer2010analysis,garcia2010pattern,gheyas2010neural}. Problems with missing data can be categorized into three types: data is completely at random (MCAR), missing at random (MAR), and not missing at random (NMAR). MCAR occurs while any feature of a data instance is missing completely random and is independent of the values of other features. Data is MAR , when the data instance with missing feature is dependent on the value of one or more of the instances’ other features. NMAR occurs when the data instance with missing feature is dependent on the value of the other missing features. Even though MCAR is more desirable, in many real-world problems, MAR occurs frequently in practice \cite{ghannad2010selection}. 

%A simple approach to deal with missing values is to ignore the data points that contain missing features and builds the classifier based on the complete data features \cite{rubin1976inference,jerez2010missing,garcia2010pattern}. This technique, commonly known as filtering, is applicable in practice, but it can be problematic in the presence of high percentage of missing features and may cause serious bias in data analysis and classification. For example in biology, ignoring the genes that consist of missing features is impractical since a large extent of data in the matrix could be discarded \cite{pan2011towards}. 

In the imputation methods, the goal is to substitute a missing value with a meaningful estimation \cite{garcia2010pattern}. This can be done  either directly from the information on the dataset or by constructing a predictive model for this purpose. Standard methods for imputation are mean imputation \cite{donders2006review}, kNN imputation \cite{batista2002study}, Bayesian principal component analysis (BPCA) imputation \cite{oba2003bayesian}, and the expectation maximization (EM) \cite{schneider2001analysis}. We apply the EM method which is  one of the most successful imputation methods \cite{huang2009maximum}. The EM method  iteratively applies linear regression analysis and fits a new linear to the estimated data until a local optimum is achieved \cite{ghahramani1994supervised,schneider2001analysis}.  In the regularized adaption of EM method, the conditional maximum likelihood estimation of regression parameters is replaced in the conventional EM algorithm \cite{nanni2012classifier}. 

\subsection{Regularized Expectation-Maximization}
In our preprocessing when the data contain many missing values, we apply the EM  algorithm. It iteratively calculates the maximum-likelihood (ML) estimates of parameters by exploring the relationship between the complete data and the incomplete data (with missing features) \cite{dempster1977maximum}. In many cases, it has been demonstrated that the EM algorithm achieves reliable global convergence, economical storage. It is not computationally expensive, and can be easily implemented \cite{redner1984mixture}. In EM we maximize the objective of the log-likelihood function
\begin{subequations}
\begin{align}  
              L(\Theta;\chi) = \sum_{i=1}^{n}log p(x_i| \Theta),
\end{align}
\end{subequations}
where $\chi = \{x_i| i=1,...,n\}$ are the observations with independent distribuation $p(x)$ parametrized by $\Theta$. 

The regularized EM algorithm (REM) is developed to control the level of uncertainty associated to missing values \cite{li2005regularized}. The main idea is to regularize the likelihood function according to the mutual relationship between the observations and the missing data with little uncertainty and maximum information. Intuitively, it is desirable to select the missing data that has a high probabilistic association with the observations, which shows that there is little uncertainty on the missing data given the observations. It performs linear regression iteratively for the imputation of missing values. The REM algorithm optimizes the penalized likelihood as follows:
\begin{subequations}
\begin{align}  
             \tilde{L}(\Theta;\chi) = L(\Theta;\chi) + \gamma P(\chi,\Upsilon| \Theta),
\end{align}
\end{subequations}
where P is the distribution function of the complete data given $\Theta$. The trade-off between the degree of regularization of the solution and the likelihood function is controlled by the so-called regularization parameter that is represented by $\gamma$ that \cite{li2005regularized}. 
In addition to reducing the uncertainty of missing data, the REM preserves the advantage of the standard EM method. This method is very efficient for over-complicated models. 

\subsection{Performance Measures}
Classification algorithms are evaluated based on the performance measures, which are calculated from the confusion matrix (\ref{wrap-tab:1}). 
\begin{table}
\caption{Confusion matrix}\label{wrap-tab:1}
\centering
\begin{tabular}{|c|c|c|}
\hline & Positive class & Negative Class\\  \hline
\multirow{2}{*}{Positive Class} & True Positive & False Positive   \\
                                  &   (TP)         &     (FP)          \\    \hline
 \multirow{2}{*}{Negative Class} & False Negative & True Negative  \\
                                &   (FN)            &    (TN)          \\
\hline 
\end{tabular}
\end{table} 
For binary classification problems, the performance measures are defined as accuracy (ACC), sensitivity (SN), specificity (SP), and G-mean, namely, 
\begin{equation}
\textrm{SN}=\frac{TP}{TP+FN}, \ \ \textrm{SP}=\frac{TN}{TN+FP}
\end{equation}
\begin{equation}
\textrm{G-mean}=\sqrt{\textrm{SP}*\textrm{SN}}
\end{equation}
\begin{equation}
 \textrm{ACC}=\frac{TP+TN}{FP+TN+TP+FN}.
\end{equation}

\section{Computational Results}

We evaluate the proposed classification framework on academic (UCI \cite{UCI}, and the cod-rna dataset \cite{alon1999broad}), and real-life binary classification benchmarks \cite{RoderickMEDAI}, \cite{RoderickFLU}. Coarsest and refinement (W)SVM models are solved using LIBSVM-3.18 \cite{chang2011libsvm}, and the FLANN library \cite{muja_flann_2009} is used to create the A$k$NN graphs. Multilevel frameworks, data processing and further scripting are implemented in MATLAB 2012a. 
The C4.5, Naive Bayes (NB), Logistic Regression (LR), and 5-Nearest Neighbor (5NN) are implemented using WEKA interfaced with MATLAB. A typical 10-fold cross validation setup is used. 
We create missing values on the academic data training sets by discarding the features randomly. The misclassification penalty or weights are selected as inversely proportional to the size of each class in our implementation. As a preprocessing step, the whole data is normalized before classification. The nested uniform design (UD) is performed on the training data as the model selection for (W)SVM \cite{huang2007model}. The UD methodology is very successful for model selection in supervised learning \cite{mangasarian2008privacy}. The close-to-optimal parameter set is achieved in an iterative nested process \cite{huang2007model}. The optimal parameter set is selected based on {\em G-mean} maximization, since data might be imbalanced. A 9- and 5-point run design is performed for the first and second stages of the nested UD due to its superiority for the UCI data \cite{huang2007model}, and the performance measures such as sensitivity, specificity, G-mean and accuracy are calculated on the testing data. 

\begin{table}[h]
\footnotesize
\centering
\caption{Academic data sets.}
\label{table1} 
    \begin{tabular}{cccccccc}         %|c|c|c|c|c|c|c|c|
    
        \hline
Dataset	&	$r_{imb}$	&	$n_f$ &	$|\JJ|$	&	$|\CC^+|$	&	$|\CC^-|$	\\ \hline
Twonorm	&	0.50	&	20	&	7400	&	3703	&	3697	\\
Letter26	&	0.96	&	16	&	20000	&	734	&	19266	\\
Ringnorm	&	0.50	&	20	&	7400	&	3664	&	3736	\\
Cod-rna	&	0.67	&	8	&	59535	&	19845	&	39690	\\
Clean (Musk)	&	0.85	&	166	&	6598	&	1017	&	5581	\\
Advertisement	&	0.86	&	1558	&	3279	&	459	&	2820	\\
Nursery	&	0.67	&	8	&	12960	&	4320	&	8640	\\
Hypothyroid	&	0.94	&	21	&	3919	&	240	&	3679	\\
Buzz  &	0.80	&	77	&	140707	&	27775	&	112932	\\
\hline
\end{tabular}
\end{table}

\subsection{Academic data sets}
We compared popular methods with the proposed ML(W)SVM to classify imperfect data.
Table \ref{table2} shows the comparative results of MLSVM, MLWSVM, SVM, WSVM, Naive Bayes, C4.5, LR, and 5NN algorithms for academic data sets. These methods are examined for different missing value ratios selected as 5\%, 10\%, 20\%, and 40\%. We implemented the REM method for missing data imputation \cite{schneider2001analysis}. The highest values are shown in boldface among all methods for their related missing value levels. It is clear from the accumulation of boldface results, MLWSVM and WSVM perform better than the other methods in general for all missing vaule ratios. In fact, MLWSVM and WSVM results into higher G-mean values in 19 out of 36 dataset/$r_{mv}$ combinations followed by MLSVM and SVM with 13 out of 36. Moreover, the ML(W)SVM techniques achieve faster computational time compared to the standard (W)SVM (Table \ref{ta2}).

%Decision trees and Naive Bayes are found to yield the best performance \cite{zhang2013impute} ({\bf TALAYEH: How do you see this? in what cases?})

% ({\bf TALAYEH: and the others \% are bad? again, how do you see this?}). 

\begin{table*}[t]
\footnotesize
\centering
\caption{Comparative G-mean results for ML(W)SVM against the regular SVM, WSVM, NB, C4.5, 5NN, and LR on academic datasets for different fractions of missing values ($r_{mv}$).}
\label{table2} 
    \begin{tabular}{cccccccccc}                  %|c|c|c|c|c|c|
        \hline
% \multirow{2}[0]{*}{Dataset}                         & \multicolumn{4}{c}{SVM}  &        \\\cline{2-6}
Dataset	&	$r_{mv}$	&	MLSVM	&	MLWSVM	&	SVM	&	WSVM	&	C4.5	&	5NN	&	NB	&	LR	\\ \hline

\multirow{4}{*}{Twonorm}	&	5\%	&	{\bf 0.98}	&	{\bf 0.98}	&	{\bf 0.98}	&	{\bf 0.98}	&	0.86	&	0.97	&	{\bf 0.98}	&	{\bf 0.98}	\\
	&	10\%	&	{\bf 0.98}	&	{\bf 0.98}	&	0.97	&	0.97	&	0.87	&	0.97	&	0.97	&	0.97	\\
	&	20\%	&	{\bf 0.98}	&	{\bf 0.98}	&	{\bf 0.98}	&	{\bf 0.98}	&	0.88	&	0.97	&	0.97	&	{\bf 0.98}	\\
	&	40\%	&	0.97	&	0.97	&	0.97	&	0.97	&	0.89	&	0.97	&	{\bf 0.98}	&	{\bf 0.98}	\\\hline
\multirow{4}{*}{Letter}	&	5\%	&	0.97	&	{\bf 1.00}	&	0.99	&	0.99	&	0.97	&	0.98	&	0.86	&	0.81	\\
	&	10\%	&	0.98	&	{\bf 1.00}	&	0.98	&	0.99	&	0.98	&	0.98	&	0.86	&	0.80	\\
	&	20\%	&	{\bf 1.00}	&	{\bf 1.00}	&	0.99	&	0.99	&	0.97	&	0.98	&	0.87	&	0.80	\\
	&	40\%	&	0.95	&	0.97	&	0.96	&	{\bf 0.99}	&	0.97	&	0.98	&	0.88	&	0.83	\\\hline
\multirow{4}{*}{Ringorm}	&	5\%	&	0.97	&	0.98	&	0.97	&	0.98	&	0.91	&	0.61	&	{\bf 0.99}	&	0.76	\\
	&	10\%	&	0.98	&	0.98	&	{\bf 0.99}	&	{\bf 0.99}	&	0.91	&	0.62	&	0.98	&	0.76	\\
	&	20\%	&	{\bf 0.98}	&	{\bf 0.98}	&	0.97	&	{\bf 0.98}	&	0.91	&	0.62	&	{\bf 0.98}	&	0.76	\\
	&	40\%	&	{\bf 0.98}	&	{\bf 0.98}	&	0.97	&	{\bf 0.98}	&	0.91	&	0.62	&	{\bf 0.98}	&	0.76	\\\hline
\multirow{4}{*}{Cod-rna}	&	5\%	&	0.95	&	{\bf 0.96}	&	{\bf 0.96}	&	{\bf 0.96}	&	0.95	&	0.92	&	0.66	&	0.93	\\
	&	10\%	&	0.95	&	{\bf 0.96}	&	0.95	&	0.96	&	0.95	&	0.91	&	0.66	&	0.92	\\
	&	20\%	&	0.95	&	{\bf 0.96}	&	0.95	&	0.95	&	0.94	&	0.91	&	0.67	&	0.92	\\
	&	40\%	&	{\bf 0.95}	&	{\bf 0.95}	&	{\bf 0.95}	&	{\bf 0.95}	&	0.93	&	0.90	&	0.68	&	0.91	\\\hline
\multirow{4}{*}{Clean}	&	5\%	&	{\bf 1.00}	&	0.99	&	0.98	&	{\bf 1.00}	&	0.83	&	0.92	&	0.79	&	0.89	\\
	&	10\%	&	0.99	&	{\bf 1.00}	&	0.99	&	{\bf 1.00}	&	0.83	&	0.91	&	0.79	&	0.89	\\
	&	20\%	&	{\bf 1.00}	&	{\bf 1.00}	&	{\bf 1.00}	&	{\bf 1.00}	&	0.83	&	0.91	&	0.79	&	0.89	\\
	&	40\%	&	{\bf 1.00}	&	{\bf 1.00}	&	{\bf 1.00}	&	{\bf 1.00}	&	0.82	&	0.92	&	0.79	&	0.89	\\\hline
\multirow{4}{*}{Advertisement}	&	5\%	&	0.87	&	0.87	&	0.87	&	0.87	&	{\bf 0.92}	&	0.81	&	0.60	&	0.82	\\
	&	10\%	&	{\bf 0.87}	&	{\bf 0.87}	&	0.86	&	0.86	&	0.86	&	0.85	&	0.62	&	0.82	\\
	&	20\%	&	0.83	&	0.85	&	0.83	&	0.85	&	{\bf 0.89}	&	0.83	&	0.61	&	0.83	\\
	&	40\%	&	0.84	&	0.86	&	0.87	&	0.81	&	{\bf 0.91}	&	0.85	&	0.62	&	0.82	\\\hline
\multirow{4}{*}{Nursery}	&	5\%	&	0.99	&	0.99	&	{\bf 1.00}	&	{\bf 1.00}	&	{\bf 1.00}	&	{\bf 1.00}	&	0.00	&	{\bf 1.00}	\\
	&	10\%	&	0.99	&	0.99	&	{\bf 1.00}	&	{\bf 1.00}	&	{\bf 1.00}	&	{\bf 1.00}	&	0.00	&	{\bf 1.00}	\\
	&	20\%	&	0.96	&	0.96	&	{\bf 1.00}	&	{\bf 1.00}	&	{\bf 1.00}	&	{\bf 1.00}	&	0.00	&	1.00	\\
	&	40\%	&	0.92	&	0.92	&	{\bf 1.00}	&	{\bf 1.00}	&	{\bf 1.00}	&	0.99	&	0.46	&	{\bf 1.00}	\\\hline
\multirow{4}{*}{Hypothyroid}	&	5\%	&	0.83	&	0.87	&	0.81	&	0.87	&	0.96	&	0.76	&	{\bf 0.97}	&	0.88	\\
	&	10\%	&	0.85	&	0.86	&	0.78	&	0.86	&	{\bf 0.96}	&	0.76	&	{\bf 0.96}	&	0.89	\\
	&	20\%	&	0.84	&	0.86	&	0.72	&	0.86	&	0.96	&	0.75	&	{\bf 0.97}	&	0.90	\\
	&	40\%	&	0.86	&	0.88	&	0.84	&	0.88	&	0.96	&	0.76	&	{\bf 0.97}	&	0.89	\\\hline
\multirow{4}{*}{Buzz}	&	5\%	&	{\bf 0.94}	&	{\bf 0.94}	&	{\bf 0.94}	&	{\bf 0.94}	&	{\bf 0.94}	&	0.93	&	0.89	&	{\bf 0.94}	\\
	&	10\%	&	{\bf 0.94}	&	{\bf 0.94}	&	{\bf 0.94}	&	{\bf 0.94}	&	{\bf 0.94}	&	0.93	&	0.89	&	{\bf 0.94}	\\
	&	20\%	&	0.92	&	{\bf 0.94}	&	0.93	&	{\bf 0.94}	&	{\bf 0.94}	&	0.93	&	0.88	&	0.93	\\
	&	40\%	&	0.93	&	0.93	&	0.93	&	0.93	&	{\bf 0.94}	&	{\bf 0.94}	&	0.86	&	{\bf 0.94}	\\
\hline
\end{tabular}
\end{table*}
\normalsize

\begin{table}[h]
\footnotesize
\centering
\caption{Computational Time ( sec.)}
\label{ta2} 
    \begin{tabular}{ccccc}                  %|c|c|c|c|c|c|
        \hline
	&	MLSVM	&	SVM	&	MLWSVM	&	WSVM	\\\hline
Twonorm	&	6	&	29	&	6	&	29	\\
Letter	&	37	&	145	&	39	&	146	\\
Ringnorm	&	5	&	26	&	5	&	27	\\
Cod-rna	&	300	&	1865	&	315	&	1891	\\
Clean	&	25	&	103	&	23	&	90	\\
Advertiesment	&	99	&	228	&	101	&	232	\\
Nursery	&	26	&	188	&	32	&	193	\\
Hypothyroid	&	2	&	3	&	2	&	3	\\
Buzz	&	3915	&	26963	&	4705	&	27732	\\

\hline
\end{tabular}
\end{table}

\subsection{Healthcare data sets}
We present the results of comparison of classification algorithms on the real-life healthcare data sets. 
Table \ref{table3} demonstrates the results on Example 1 (see Section \ref{sec:intro}), a classification task of assigning a patient in a correct group by financial risk, which are ordered in ascending manner from group 1 with the lowest level of risk, to group 5 with the highest level of risk.

The motivation behind the original study was to determine how much integration of the medical and financial data changes the outcomes of clustering and classification operations based on financial data alone. For that purpose, modest precision was sufficient; we used a logistic (linear) regression (LR) approach (implemented as \textit{mnrfit} in MATLAB). We are comparing the accuracy of it with the best results obtained by ML(W)SVM. The strategy "one-against-all" is used for multi-class classification. This strategy performs training a classifier per class with the data points of that class as positive class and the rest of the data points are trained as negative class. 

% {\bf TALAYEH: can you explain what are these numbers? ascending, descending order? something else?})

\begin{table}[h]
\caption{Accuracy of financial risk problem with five risk classes (Example 1)}
\label{table3}
\centering
\begin{tabular}{cccccc}
	\hline Class  & 1 & 2 & 3 & 4 & 5 \\ \hline
	 LR & 0.58 & 0.54 & 0.53 & 0.51 & 0.59 \\ 
	 MLSVM & 0.73 & 0.50 & 0.44 & 0.50 & 0.71 \\ 
	\hline 
\end{tabular} 
\end{table}

To interpret the results, we note that correct identification of intermediate risk categories is a very difficult problem in medical informatics. To our knowledge, there is no good definition of "average health", either evidence-driven or philosophical, that would help an expert to identify such patient features that do not indicate an acute crisis, or an almost certain safety from crisis. Accordingly, it is not surprising that neither approach does well on the risk categories 2..4; there is also not a lot of motivation to improve the model there.
On the other hand, it is important to identify and predict the very low-risk patients (knowing that status ahead of time allows resource re-allocation leading to savings and improved service for everyone) and the very high-risk patients (so that clinical and financial resources could be prepared for the forthcoming crisis).

Accordingly, it is important that the use of an advanced method of machine learning changes the quality of prediction from almost worthless ('toss a coin') to workable (accuracy of $0.7$). 

In Table \ref{table4}, we compare results for the widely used basic approach and ML(W)SVM prediction for Example 2 (see Section \ref{sec:intro}), a study of patient's response to hospital flu outreach. In this problem, the goal is to find a binary classifier that will predict whether the patient will get vaccinated after reminder, or not (this includes using a different provider for vaccination). In the preliminary study, we used adaptive linear regression model (LASSO for adaptive selection of features, logistic regression on actual prediction). 

\begin{table}[h]
\caption {Comparison of Multilevel WSVM against Multilevel SVM and Adaptive Logistic Regression (LR). Improved results are in bold.}
\label{table4}
\centering
\begin{tabular}{ccccc}
	\hline &G-mean & SN   & SP & ACC \\ 
	\hline Adaptive LR& 0.7516  & 0.8903 & 0.6345 & 0.7619 \\ 
	 MLSVM & {\bf 0.8012}   & {\bf 0.9750} & {\bf 0.6583} & {\bf 0.8496} \\ 
	 MLWSVM & {\bf 0.8016}     & {\bf 0.9739} & {\bf 0.6598} & {\bf 0.8495} \\ 
	\hline 
\end{tabular} 
\end{table}

Response to outreach is not a crucial life-or-death issue, we are performing this study to see if predictive modeling can assist with resource allocation (which patients to contact, how much medical personnel effort to dedicate to outreach and then vaccination). Arguably, accuracy is more important than specificity here. 
Even the basic results (using linear regression) were met with approval the CPSL (Care Patient Service Line: a division responsible for coordinating efforts of local, small-scale healthcare providers operating under Geisinger). SVM methods (almost 10 percent improvement) provide additional justification for the use of machine learning on merged data to assist planning in clinical practice.

\section{Conclusion}

Large-scale data, missing or imperfect features, skewness distribution of classes are common challenges in pattern recognition of many healthcare problems. We have successfully extended a powerful machine learning technique, support vector machines, to the \emph{scalable multilevel framework} of cost-sensitive learning SVM to deal with imbalanced classification problems. Our multilevel framework substantially improves the computational time without losing the quality of classifiers for large-scale datasets. We have shown that MLWSVM produces superior results than MLSVM and the regular SVM methods in most cases. This work can be extended to tackle other classification problems with large-scale imbalanced data (combined from different sources) with missing features in healthcare and engineering applications.

From the perspective of evidence-driven healthcare, our work shows that application of cutting edge machine learning techniques (in this case, fast multilevel classifiers) makes enough of a difference to justify the additional development effort for typical examples from clinical practice. While the improvements in precision and specificity we show in this study are both under $10 \%$ and are modest in general perspective, the result in healthcare is significant.

To our knowledge, such complex combined behavioral/operational phenomena as inference of financial risk from medical history (Example 1), or prediction of effectiveness of public outreach (Example 2), don't have a satisfactory casual explanation. The classical (1990s) clinical practice offered two equally unsatisfactory options: not having a capability for prediction at all, or relying on very basic statistical techniques (based on a single data source, with very high rate of false-positive classification outcomes). The existing mature models (such as actuarial projections of financial risk) do not benefit from integration of data from multiple sources, and may, in fact, turn out to be ineffective outside of their scope in patient population and metrics of interest (as we have shown in \cite{RoderickMEDAI}). Thus, in the modern clinical practice we have to rely on newly developed machine learning tools, tuned on data from multiple sources.
Thus, our work can also be extended to handle other classification problems on massive, multi-format medical data.

%=========================================================
\bibliographystyle{plain}
\bibliography{roderick_medical,paper,ilya}
\end{document}